\documentclass{article}
\usepackage{iclr2017_conference, times}
\usepackage{amsmath}
\usepackage{graphicx}
\usepackage{color}
\usepackage{multirow}
\usepackage{rotating}
\usepackage{amssymb}
\usepackage{latexsym}
\usepackage{amsmath}
\def\permille{\ensuremath{{}^\text{o}\mkern-5mu/\mkern-3mu_\text{oo}}}

\newcommand{\reviewed}[1]{{\color{black} #1}}

\usepackage{caption}
\usepackage{hyperref}
\hypersetup{
    colorlinks,
    linkcolor={red!50!black},
    citecolor={blue!50!black},
    urlcolor={blue!80!black}
}
\usepackage{url}
\usepackage{subcaption}
\usepackage{tikz}
\usetikzlibrary{bayesnet, patterns, decorations.pathreplacing}
\newcommand{\lb}{\\[6pt]}                              
\newcommand{\eq}[1]{\begin{align}{#1}\end{align}}    

\newcommand{\mom}[1]{\left\langle{#1}\right\rangle}    
\newcommand{\wvec}{\left[\hspace{-4pt}\begin{array}{c} {\mathbf{a}^{(1)}}^\intercal \\[2pt] {\mathbf{a}^{(2)}}^\intercal	\end{array}\hspace{-4pt}\right]} 
\newcommand{\wvectwo}{\left[\hspace{-4pt}\begin{array}{c} {\mathbf{a}^{(2)}}^\intercal \\[2pt] {\mathbf{a}^{(1)}}^\intercal	\end{array}\hspace{-4pt}\right]} %

\title{Multi-View Bayesian Correlated \newline Component Analysis}

\author{Simon Kamronn$^{\ast}$, Andreas Trier Poulsen\thanks{Equal contribution\newline This is a pre-print of Simon Kamronn, Andreas Trier Poulsen, and Lars Kai Hansen. Multiview Bayesian Correlated Component Analysis. \textit{Neural Computation}, 27, (10):2207–30, 2015} \ \& Lars Kai Hansen \\
Department of Applied Mathematics and Computer Science\\
Technical University of Denmark\\
Kongens Lyngby\\
Denmark\\
\texttt{\{sdka, atpo, lkai\}@dtu.dk}
}

\iclrfinalcopy
\begin{document}

\maketitle

\begin{abstract}
  Correlated component analysis as proposed by Dmochowski et al.\ (2012) is a tool for investigating brain process similarity in the responses to multiple views of a given stimulus. Correlated components are identified under the assumption that the involved spatial networks are identical. Here we propose a hierarchical probabilistic model that can infer the level of universality in such multi-view data, from completely unrelated representations, corresponding to canonical correlation analysis, to identical representations as in correlated component analysis. This new model, which we denote Bayesian correlated component analysis, evaluates favourably against three relevant algorithms in simulated data. A well-established benchmark EEG dataset is used to further validate the new model and  infer the variability of spatial representations across multiple subjects.
\end{abstract}


\section{Introduction}
The large scale organization  of information processing in the human brain is \reviewed{currently of high} interest, see e.g.\ \cite{Bullmore2009}. In particular, we are interested in the spatio-temporal networks involved when the human brain solves computational problems such as decoding high level information in digital media, say in a movie.  As we may assume that the movie brain interaction is jointly optimized for the viewing process, it seems natural to hypothesize a certain amount of universality in the representations and activation patterns  in the brains of subjects  watching a given movie.
Such  ecologically inspired neuroscience, said to be based on naturalistic stimuli, has been pursued by Hasson et al.\ (see e.g. \cite{Hasson2004}). They introduced a correlation approach between anatomically aligned brains. This is based in a rather strong assumption of representational universality, namely that both the extracted information (what) and  spatial networks  (where) are shared among subjects.
To exploit the full spatio-temporal patterns of correlation and increase sensitivity, a multivariate version of this approach, so-called correlated component analysis (here abbreviated as CorrCA) was recently proposed by  \cite{Dmochowski2012}. The method was further developed and applied in \cite{Dmochowski2014}. 
Within the multivariate framework, a natural relaxation of the strong universality assumption, is to hypothesise that decoded content, i.e, the 'what', is identical and shared between subjects, while the spatial representations can be formed individually. Such an approach corresponds to the multivariate technique known as canonical correlation analysis (CCA) \citep{Hotelling1936}. In  CCA we search for individual stationary spatial networks with similar temporal activation among subjects. CCA was generalized to account for both joint and individual signal components by \cite{Lukic}.

Inspired by probabilistic principal component analysis introduced by \cite{Tipping1999}, a probabilistic approach to CCA was presented in \cite{Bach2005}. To avoid discrete model selection in probabilistic PCA, \cite{Bishop1999} invoked the automatic relevance determination mechanism (ARD, \cite{Hansen1994, MacKay1996}), where each component, corresponding to a column in the mixing matrix, has a Gaussian distribution whose width is controlled by a gamma distributed hyperparameter, potentially shrinking to zero and effectively pruning a component and thereby automatically inferring the number of components.
These schemes for probabilistic PCA and CCA has lead to Bayesian CCA \citep{Wang2007, Wu2011, Klami2013} and so-called group factor analysis (GFA) \citep{Klami2014}, the first practical multi-view generalization of Bayesian CCA.  These methods differ in the way they approximate the hidden source posterior distribution, typically applying variational inference schemes, as is also the case for the work presented here.

\reviewed{With the work of Dmochowski et al. on EEG analysis as an inspiration, we will analyse a probabilistic model which focuses on extracting joint components, however, with the possibility of learning the degree of universality from data.}
The latter is implemented by a hierarchical Bayesian model that allows for variable non-universality in the representations used by the views, i.e., individual subjects. \reviewed{Probabilistic approaches to modelling dependencies between the representations of different views have earlier been presented in \cite{Lahti2009} and more recently in \cite{Klami2014}. In contrast to the latter, which seeks to model dependencies between views and components through the precision of the representations, the model presented here assume an average pattern and lets the precisions decide the degree of deviation from the common representation. This enables symmetry amongst the views as in CorrCA, unlike the model presented in \cite{Lahti2009}.}

We illustrate the performance of our approximate inference procedures in both simulation studies and in a benchmark EEG set.
The specific contributions of this work are: 1)  Formulation of a generative model and inference for Bayesian correlated component analysis (BCorrCa); 2) A principled scheme for inference of correlated components for more than two simultaneous views; 3) Validation on simulated data and benchmark EEG data.

\noindent\textbf{Notation}\\
Matrices are denoted by capital bold symbols, e.g. \textbf{A} \(\in \mathbb{R}^{I \times J}\), and vectors as lowercase bold symbols, e.g. \textbf{a}. The vector of a matrix is indexed as a row vector, $ \textbf{a}_i $, or column vector, $ \textbf{a}_j $, and the elements as $ a_{ij}$. To be able to distinguish row from column vectors, the lower case index letter match the upper case letter used when defining the corresponding matrix. In the case of multiple views each view is indexed as e.g. $ \textbf{A}^{(m)} $ for $m \in \{1,...,M\}$. \textbf{I} and \textbf{0} denote the identity and all-zero vector or matrix respectively, depending on context.

\section{The model}
\begin{figure}
\centering
	\begin{tikzpicture}
      
      
        \node[obs]          (x)   {$x^{(m)}_{dn}$}; %
        \factor[above=of x] {x-f} {right:$\mathcal{N}$} {} {} ; %
      
        \node[det, above=of x, xshift=0.8]         (dot) 	{dot} ; %
        \node[latent, above left=2 of dot]  		(w)   	{$\mathbf{a}^{(m)}_{dk}$}; %
        \node[latent, above right=2.5 of dot] 		(z)   	{$\mathbf{z}_{kn}$}; %
        \node[latent, above=1 of w, xshift=-1.5cm]	(l)	{$ \lambda $};
        \node[latent, above=1 of w, xshift=0.8cm]	(v)	{$ \mathbf{u}_{dk} $};
        \node[latent, above=1 of v, xshift=0.5cm]	(a)	{$\alpha_k$};
      
        \node[const, above=1.2 of v, xshift=-0.5cm] (mv) 	{$0$} ; %
      
        \node[const, above=1.2 of z, xshift=-0.5cm]	(mx) {$0$} ; %
        \node[const, above=1.2 of z, xshift=0.5cm]	(ax) {$\mathbf{I}$} ; %
        
        \node[const, above=1.2 of a, xshift=-0.5cm] (a_ard) 	{$a_\alpha$} ; %
        \node[const, above=1.2 of a, xshift=0.5cm] 	(b_ard) 	{$b_\alpha$} ; %

       \node[const, above=1.2 of l, xshift=-0.5cm] 	(la_ard) 	{$a_l$} ; %
       \node[const, above=1.2 of l, xshift=0.5cm] 	(lb_ard) 	{$b_l$} ; %
      
        \node[latent, left=3cm of x-f]         		(p)   {$\Psi^{(m)}$}; %
        \node[const, above=of p, xshift=-0.5cm] 	(at1)  {$S$} ; %
        \node[const, above=of p, xshift=0.5cm]  	(bt1)  {$v$} ; %

        \factor[above=of z] {z-f} {left:$\mathcal{N}$} {mx,ax} {z} ; %
                
        \factor[above=of a] {a-f} {left:$\mathcal{G}$} {a_ard,b_ard} {a};
        \factor[above=0.5 of l] {l-f} {left:$\mathcal{G}$} {la_ard,lb_ard} {l};
        \factor[above=0.5 of v] {v-f} {left:$\mathcal{N}$} {mv,a} {v} ; %
        \factor[above=of w]	{w-f} {left:$\mathcal{N}$} {v,l} {w};
        \factor[above=of p] {p-f} {left:$\mathcal{W}$} {at1,bt1} {p} ; %
        \factoredge {dot,p} {x-f} {x} ; %

        \edge[-] {w,z} {dot} ;
      
        \plate{wza}{
        (w)(z)(a)(ax)(a_ard)(mv)
        }{$K$};
              
        \plate {xz} { %
          (x)(z)(dot)(ax)(wza.east)
          } {$N$} ;         
        
        \plate {xw} {%
          (x)(w)(xz.north)(xz.south)(wza.west)
        } {$D$} ;
        
        \plate{xwp}{
        (p)(x)(w)(xw.south east)(at1)
        }{$M$};

      \end{tikzpicture}
  \caption{Factor graph illustrating the relationship between the variables in BCorrCA.}
	\label{fig.factorgraph}
\end{figure}
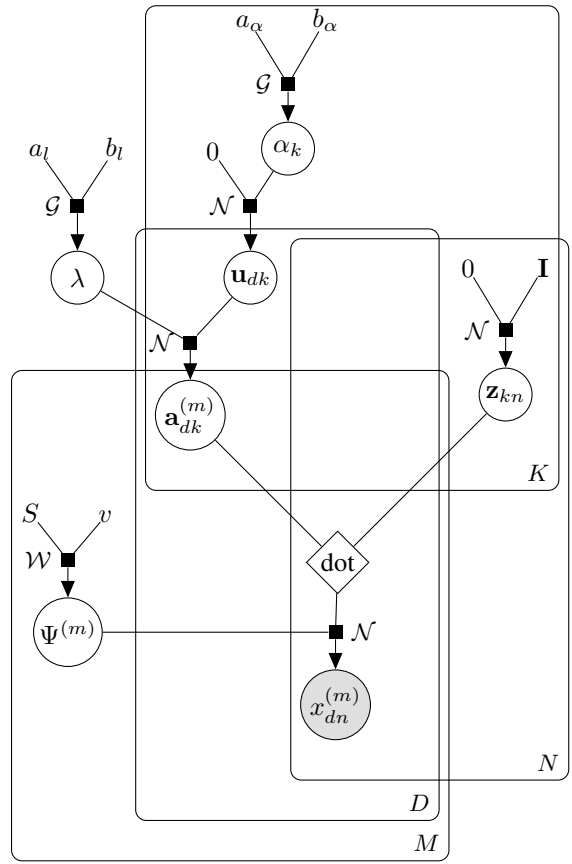
The objective of the model is to infer the decomposition of measured neural activity into shared maximally correlated time series and corresponding individual patterns as depicted in the factor graph Fig.\ \ref{fig.factorgraph}. Thus, we consider the normally distributed generative model

\begin{align}
\textbf{X}^{(m)} &\sim \mathcal{N}\left(\textbf{A}^{(m)} \mathbf{Z}, {\mathbf{\Psi}^{(m)}}^{-1}\right)\\
\textbf{Z} &\sim \mathcal{N}(\textbf{0}, \textbf{I}),
\end{align}
where $ \textbf{X}^{(m)} \in \mathbb{R}^{D \times N} $ denotes the observed neural activity for each view \textit{m}, $ \textbf{A}^{(m)} \in \mathbb{R}^{D \times K} $ is the forward model 'pattern' (\cite{Biessmann2012}), $ \textbf{Z} \in \mathbb{R}^{K \times N} $ denotes the set of shared latent sources, and $ \mathbf{\Psi}^{(m)} $ is the precision. Using probabilistic formulations it is straightforward to extend dimensionality of variables to facilitate e.g. group-wise analysis with multiple views. However, this formulation is common to a variety of models, including Bayesian CCA, so the way the models differentiate, is in the higher levels of the hierarchy. Specifically, the novelty of BCorrCA lies in the relationship between the individual patterns

\eq{
\textbf{A} \sim \prod_m^M\prod_k^K\mathcal{N}\left(\textbf{a}_k^{(m)}|\textbf{u}_k, \lambda^{(m)^{-1}}_{k}\right), \quad \textnormal{for } M\geq 2,
}
where each column of $ \textbf{A}^{(m)} $ is a pattern corresponding to one component. The latent variable,

\eq{
\textbf{U} \sim \prod_k^K\mathcal{N}\left(\textbf{u}_k|\textbf{0}, \alpha^{-1}_k\right),
}
represents the common pattern across all views and the precision $\lambda$ regularizes the distance between the view-specific patterns $\textbf{A}^{(m)}$ and \textbf{U}. With prior knowledge of the number of hidden sources, \textit{K}, the dimension of the estimated forward model can be reduced, which is an advantage when $ K \ll D $ and \textit{D} is large.
\subsection{Priors on the noise}
Since the factorisation $ \textbf{X} \simeq \textbf{AZ} $ is not expected to adequately explain the noise covariance of the neural activity, the model includes a full \reviewed{rank} precision matrix. \reviewed{This choice is motivated by the large number of sources contributing to the noise signal and by volume conduction leading to inter-channel correlation \citep{Wipf2010}.}

\reviewed{For convenience, the precision is assigned a Wishart prior,}
\eq{
 \boldsymbol\Psi^{(m)} \sim \mathcal{W}(\textbf{S}_0,v_0),
}
where $ \textbf{S}_0 $ and $ v_0 $ are hyperparameters.  The prior on the precisions of the shared mixing matrix, \textbf{U}, is likewise the conjugate gamma distribution;
\eq{
\alpha_k \sim \mathcal{G}a(a_0,b_0),
}
where $ a_0 $ and $ b_0 $ are shape and scale parameters. If the hyperprior is 'flat', the hyperparameters of the prior are estimated from the data and if the number of features are larger than the dimension of the latent sources, the prior on the precision will assume large values for some components, forcing the pattern coefficients to zero. This  is the called automatic relevance determination (ARD) mechanism and is widely used in Bayesian models to prune irrelevant features \citep{Hansen1994,MacKay1996}. The prior on the precision for individual patterns, \textbf{A}, can also be considered an ARD type parameter and is also modelled with the gamma distribution
\eq{
\lambda^{(m)}_{k} \sim \mathcal{G}a(a_0,b_0).
}
The model has the ability to share the precision hyperprior across both views and patterns to enforce shared representation or to be inferred individually for maximum flexibility. For simplicity  $ \lambda $ is assumed shared in the following.

\subsection{Inference}
The posterior distribution of all the latent, or hidden, variables \textbf{H},

\eq{
p(\textbf{H}|\textbf{X}) = \frac{p(\textbf{H},\textbf{X})}{p(\textbf{X})},
}
is analytically intractable. In probabilistic variational inference we invoke a simplifying assumption, namely that the approximate posterior distribution, \textit{q}, is completely factorised as

\eq{
q(\textbf{H}) = \prod_i q_i(\textbf{H}_i).
}
 Thus we  obtain a posterior distribution for each factor separately. This simplification is originally known in physics as \textit{mean field theory} \citep{Bishop2006}. The BCorrCA model presented in this article is based on variational inference with assumptions of a factorised posterior and conjugate priors from the exponential family. An analytic solution is attained for each variable, which are then updated iteratively in an expectation-maximisation like manner until convergence. The updates for BCorrCA are provided in detail in the appendix. The result is an algorithm, which can implement both independent mixing  as in CCA (with a small $\lambda$), or completely aligned matrices as in CorrCA (with a large $\lambda$). Importantly  it generalises in a straightforward manner to an arbitrary number of parallel data views.

A lower bound function, $\mathcal{L}(q)$, for the full log marginal likelihood is often calculated to monitor convergence. It is usually derived as the sum of the expectations of all prior and posterior distributions calculated independently. Inspired by \cite{Murphy2012} we have chosen to combine the expectations into one equation, replace terms with variables already calculated (such as $b_\lambda$), and let them cancel each other out, where applicable. Since it is the change of the lower bound that is of interest, we also combined all constant terms into the common constant, $C$. Compared to \cite{Wu2011} and \cite{Wang2007} which present the lower bound for similar models\footnote{The lower bound for the model presented in \cite{Wang2007} appears in the R-code supplied for the paper}
the result is a more compact and cost efficient calculation

\eq{
\mathcal{L}(q) &= \frac{1}{2} \sum_m^M\left\{ v_\Psi \ln|S_\Psi^{(m)}| + \sum^D_d \ln|\Sigma_{\textbf{a}_d}^{(m)}| \right\} -a_\lambda\ln b_\lambda\notag\\
&\quad + \sum^K_k \left\{ -a_\alpha \ln b_{\alpha_k} + \frac{D}{2}\ln\sigma^2_{\textbf{u}_k} \right\}
 + \frac{N}{2}\ln |\Sigma_\textbf{z}| \notag\\
& \quad - \frac{1}{2}\left( N\cdot\textnormal{Tr}(\Sigma_\textbf{z}) + \sum^N_n\mu_{\textbf{z},n}^T\mu_{\textbf{z},n}  \right) + C.
} 
\subsection{Selecting hyperparameters for prior distributions}
In a fully Bayesian setting the hyperparameters are often set to make the priors as non-informative as possible so that they will be estimated from data. The default non-informative settings in this model for the Wishart distribution are $ \mathbf{S_0} = 10^{-3}\textbf{I} $ and $ v_0 = D+1 $ and for both gamma distributions are $ a_0 = b_0 = 10^{-3} $. \reviewed{ These settings ensures a high level flexibility in the prior which can regularise even high dimensional data.} If \textit{a priori} knowledge of the system is available the hyperparameters can be changed to reflect that or an empirical Bayesian approach can be applied by estimating the hyperparameters from data. By e.g. using the pre-stimulus EEG in an ERP paradigm or baseline recordings, it is possible to estimate measurement variance which can be used to set hyperparameters of $ \boldsymbol{\Psi} $, \reviewed{which is applied in the EEG experiment in section \ref{sec:EEG}.} 
\section{Performance on simulated data}
To validate  BCorrCA and quantify its performance  data was generated with a varying similarity between the mixing matrices by changing the 'true' $\lambda$. From the model definition we get that $	\mathbf{X}^{(m)} = \textbf{A}^{(m)}_{\text{true}}\mathbf{Z} + \boldsymbol{\epsilon}$ with $ \boldsymbol{\epsilon} \sim \mathcal{N}(0,\sigma^2_\epsilon) $, where $\sigma^2_\epsilon$ is varied to obtain a desired signal-to-noise ratio (SNR) \reviewed{and the dimensionality of the oberservation space is set to $D=6$}. \textbf{Z} is a $K\times N$ source matrix containing \textit{K} time series\reviewed{, where we apply $K=K_0$ to simplify the comparisons}. The mixing matrix is generated by $\textbf{A}^{(m)}_{\text{true}} = \mathbf{U}_{\text{true}} + \boldsymbol{\delta}^{(m)}$,  with $\textbf{U}_{\text{true}} \sim \mathcal{N}(0,\reviewed{1})$ and $\boldsymbol{\delta}^{(m)} \sim \mathcal{N}(0,\lambda^{-1}) $. \reviewed{Note that we use the simpler the noise model of GFA for fair comparison.}

We have used up to \reviewed{$K_0 = 4$ hidden sources}, generated as in  \cite{Klami2013}, for direct comparability with this work. Here we will mainly focus on the simple case of one hidden source corresponding to the data being generated from one sinusoid and additive noise.
For comparative analysis  the performances of BCorrCA, CorrCA, CCA, and GFA were estimated on the same data. For each combination of conditions either 20 or 100 datasets were randomly generated and the average correlation coefficient between the inferred source and the true source was chosen as the measure of performance.
The algorithms were tested at varying levels of SNR, number of views, $M$, and similarity between the true mixing matrices of each view. In each test the data had six dimensions and a total of 5,000 samples divided equally among the views.
CorrCA and CCA are designed to handle two views at a time. In case of multiple views  we employed the scheme for combination proposed in \cite{Dmochowski2012}, i.e., the views are concatenated in time so that all pairwise combinations are compared. This method has the disadvantage that the number of samples in the concatenated data scales with the number of views as $M(M-1)$.

\subsection{Infering the similarity between mixing matrices}
\begin{figure}[ht]
\centering
	\includegraphics[width=.8\linewidth]{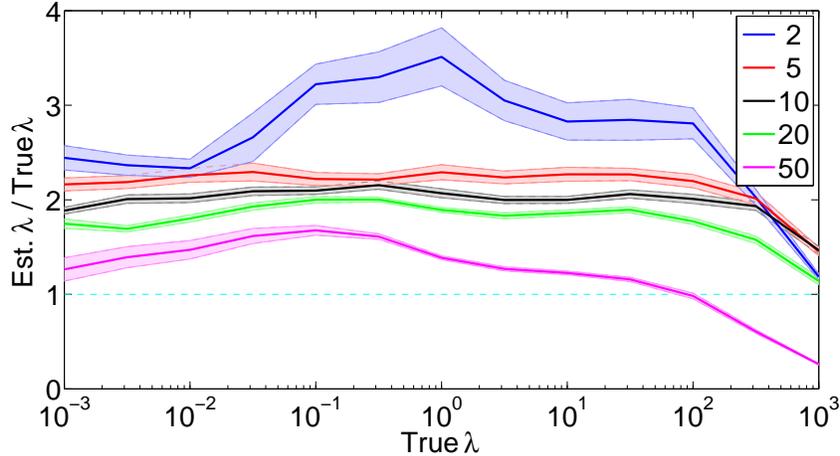}
	\caption{Estimation of the similarity between the mixing matrices in simulated data with different number of views. The opaque area marks the standard error of the mean. The similarity is regulated through the parameter $\lambda$, and estimated using BCorrCA. The simulations were conducted with a single hidden source, 100 repetitions and a SNR of 3 dB.}
	\label{fig.varL_iL}
\end{figure}
Figure \ref{fig.varL_iL} shows the results of simulations where the similarity between the true mixing matrices is varied by the $\lambda$ parameter. It can be seen that BCorrCA is able to estimate this parameter through the entire range, though it has a tendency to overestimate. This overestimation might stem from an interaction with the other ARD-parameter, $\alpha$, since the magnitude of the weights in the mixing matrix is influenced by both the $\alpha$ and $\lambda$ parameters. The interaction might also be the cause of the drop seen when the true value of $\lambda$ is larger than $10^2$. As both $\alpha$ and $\lambda$ are initialised at 1, it is possible that BCorrCA finds a local optimum closer to the initialised value of $\lambda$ and compensates through $\alpha$.
Finally it can be seen that increasing the number of views improves the precision. An effect that is most pronounced when dealing with a single hidden source and lower noise levels.
\subsection{Using the lower bound as a performance measure}
\begin{figure}
	\centering
		\includegraphics[width=\linewidth]{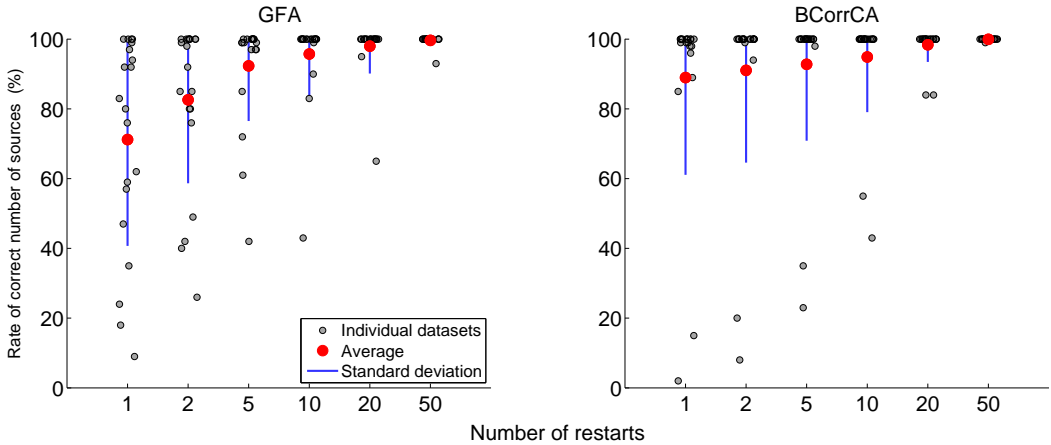}
		\caption{\reviewed{Relationship between the number of random restarts and obtaining the correct number of sources. 20 datasets were randomly generated with the same parameters; $M=5$ views, $K_0=4$ true sources, $\textnormal{SNR}=-3$ dB, $\lambda = 1$ mixing matrix similarity, and $D=8$ dimensionality of observation space. The algorithms were initialised with $K=6$ hidden sources in 100 analyses on each dataset, in which the lower bound was used to select amongst a varying number of restarts. Each grey dot shows the accuracy of choosing the right number of active sources, with the red dot and blue line signifying the grand average and standard deviation.}}
		\label{fig.as_proc}
	\end{figure}
While CCA and CorrCA are deterministic, the latent variable models are quite dependent on their (random) initialisation. This creates a need for a way to choose the best amongst multiple analyses on the same dataset. Inspired by the use of the lower bound as performance indicator in \cite{Klami2013}, \reviewed{we investigated the relationship between the calculated lower bound and the ability to infer the correct source space. To do this we randomly restarted BCorrCA and GFA 100 times on each of 200 datasets, with varying mixing matrix similarity $\left(\lambda = \{10^{-3}, 1, 10^{3}\}\right)$ and dimension of the observation space $\left(D=\{8,50\}\right)$.

For each dataset, the lower bound and the accuracy of inferring the true sources\footnote{Measured as the correlation between inferred and true sources.} was correlated across the random restarts. BCorrCA and GFA obtained an average correlation of -0.02 and -0.03 suggesting that there is no consistent relationship between the lower bound and inferring the correct source space. However, with standard deviations of 0.22 and 0.19 for BCorrCA and GFA, the relationship does seem to be dataset dependent. A trend that was seen in all combinations on dataset parameters.}

\reviewed{When correlating the true and inferred sources, each true source was paired with one inferred source without replacement, such that the largest possible average correlation was obtained. This approach does not penalise the algorithm if it infers too many sources. We therefore decided to test if the lower bound is a good indicator of whether the algorithms achieve the right number of active sources through ARD. A source was defined as inactive when its reconstructed variance was below 1\permille\ compared to the source with the highest variance. The reconstructed variance was defined as the product of the source variance and the variance of the corresponding inferred pattern.

The test showed, that there was a clear correlation between lower bound and obtaining the true number of active sources, prompting an investigation into how many restarts are necessary before one can be sure that the correct number of sources are obtained. Figure \ref{fig.as_proc} shows the relationship between the number of restarts on the same dataset and the accuracy of inferring the right number of sources. In most cases a few restarts are sufficient for both algorithms, but some datasets, though generated with the same parameters, prove more difficult and needs more restarts to ensure the algorithm has converged on the correct number of sources. The number of restarts necessary varies with the dataset parameters, some combinations proving more difficult while others enabled the algorithms to always converge on the correct number of sources.

In the special case of datasets with low dimensionality and equal representations GFA produced too few active sources (data not shown). Under these conditions ($D=8, \lambda=10^3$) GFA produced three active sources on 21 out of 40 datasets, which contained four true sources. In 8 of these 21 datasets too few sources resulted in a higher lower bound. Changing either dataset parameter meant that GFA only obtained the right number of sources or more. We were not able to make BCorrCA obtain too few sources.}

\reviewed{Initial tests showed that BCorrCA had a tendency of producing too many active sources, when the Wishart prior for the noise had been uninformed $\left(\Psi^{(m)} \sim \mathcal{W}\{10^{-3}\cdot I,v_0\}\right)$. Running the same tests using the variance of the observations $\left(\Psi^{(m)} \sim \mathcal{W}\{\text{var}[X^{(m)}]\cdot I,v_0\}\right)$ alleviated this problem and BCorrCA was able to obtain the correct number of sources. For the sake of fair comparisons the same approach was tried for the gamma prior used in GFA $\left(\tau^{(m)} \sim \mathcal{G}a\{1,\text{var}[X^{(m)}]\} \right)$. This, however, did not results in an improvement, probably due to the simpler noise model.}

\subsection{Performance under varying conditions}
\begin{figure*}[!htb]
	\centering
	\begin{subfigure}[b]{0.48\linewidth}
		\includegraphics[width=\textwidth]{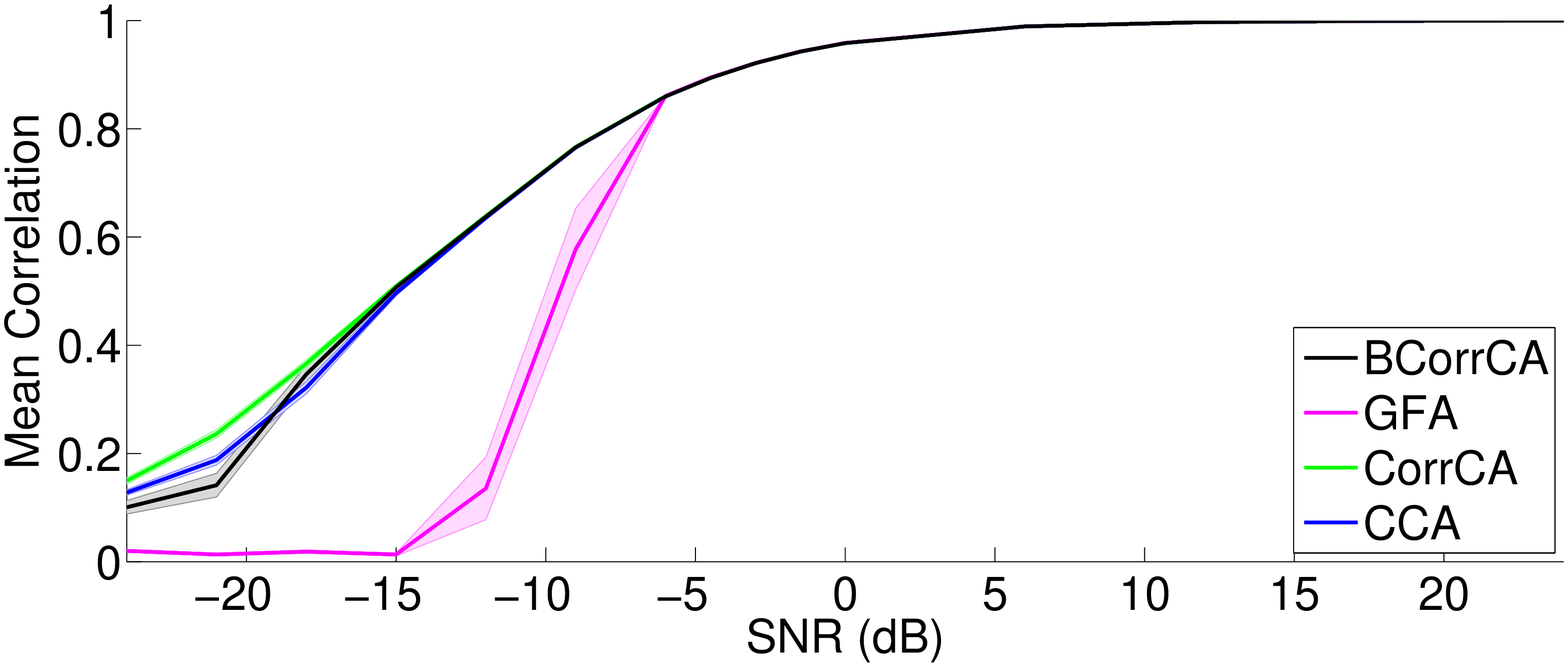}
		\caption{Two views}\label{fig.simsnr1}
	\end{subfigure}
	\hspace{2pt}
	\begin{subfigure}[b]{0.48\linewidth}
		\includegraphics[width=\textwidth]{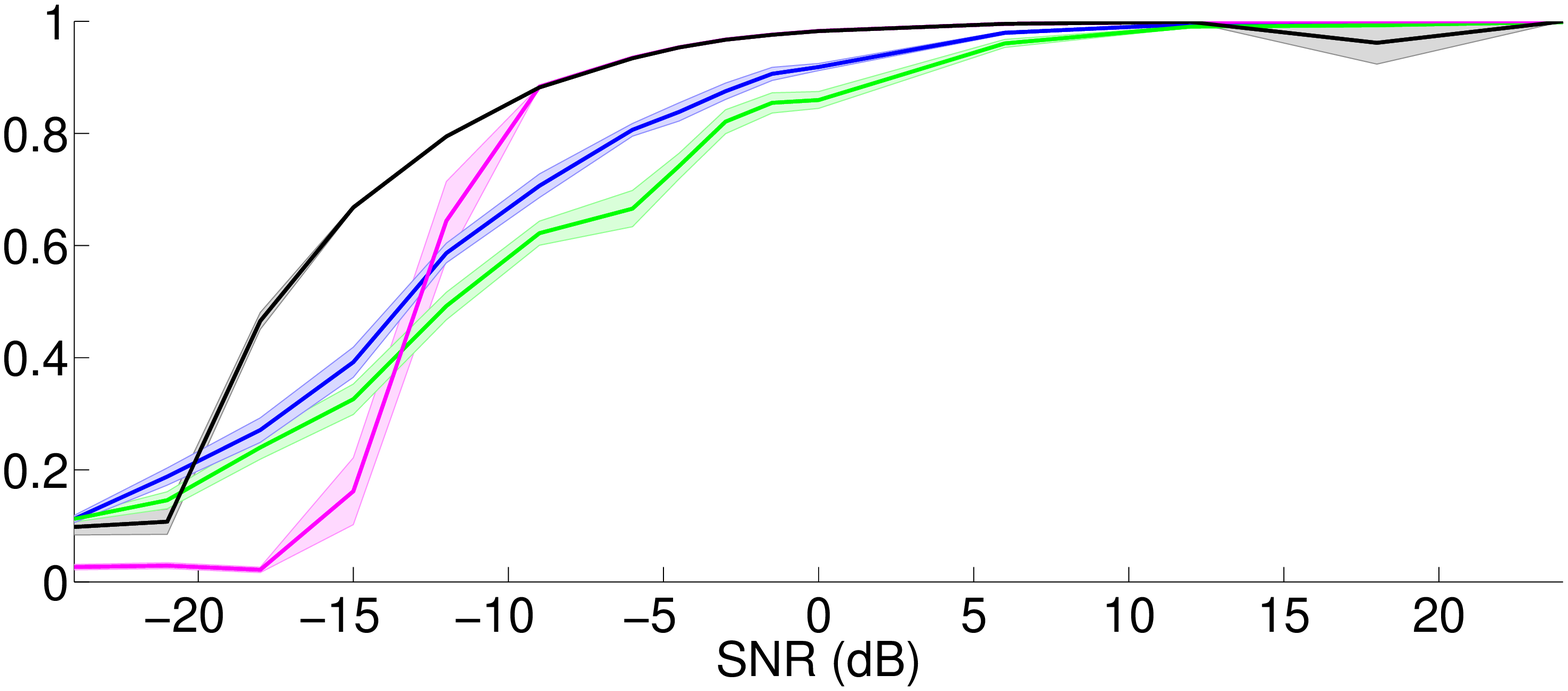}
		\caption{Five views}\label{fig.simsnr4}
	\end{subfigure}\\
	\begin{subfigure}[b]{0.48\linewidth}
		\includegraphics[width=\textwidth]{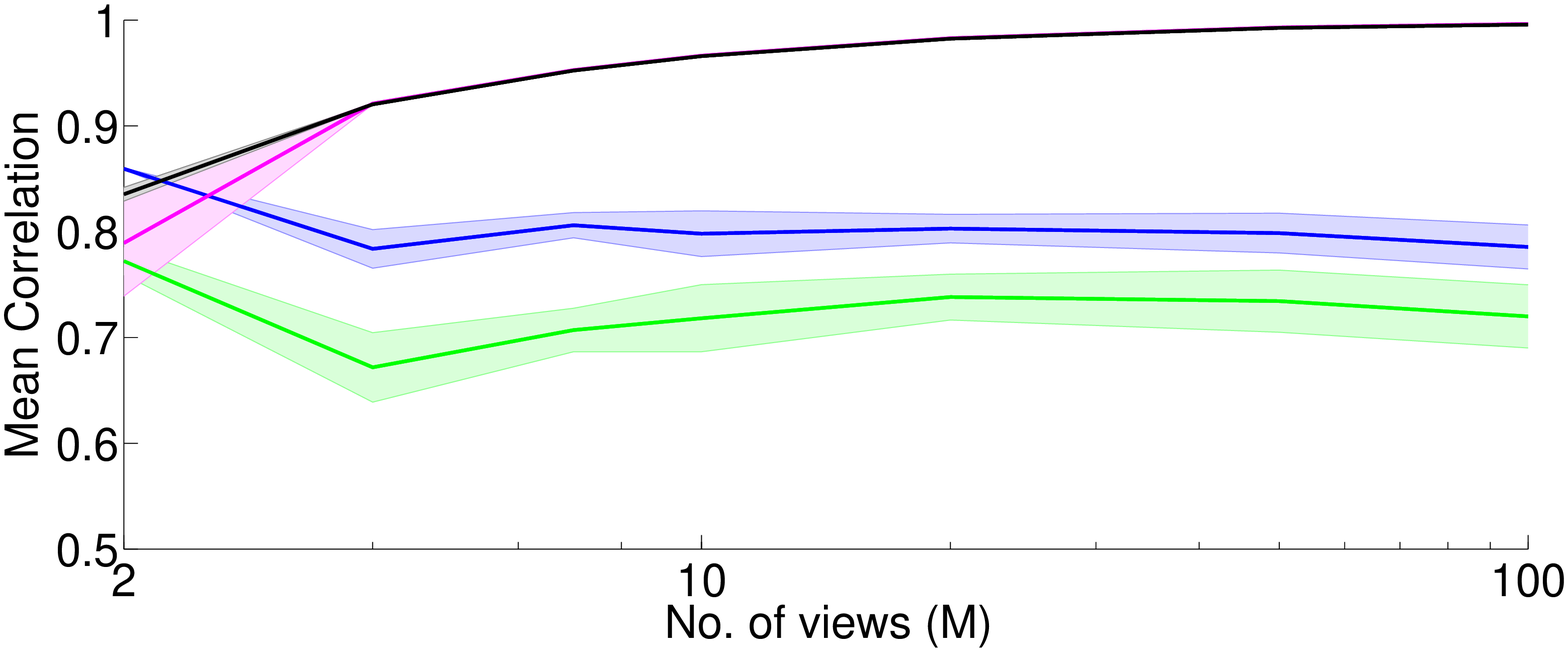}
		\caption{Varying M}\label{fig.simm}
	\end{subfigure}
	\hspace{2pt}
	\begin{subfigure}[b]{0.48\linewidth}
		\includegraphics[width=\textwidth]{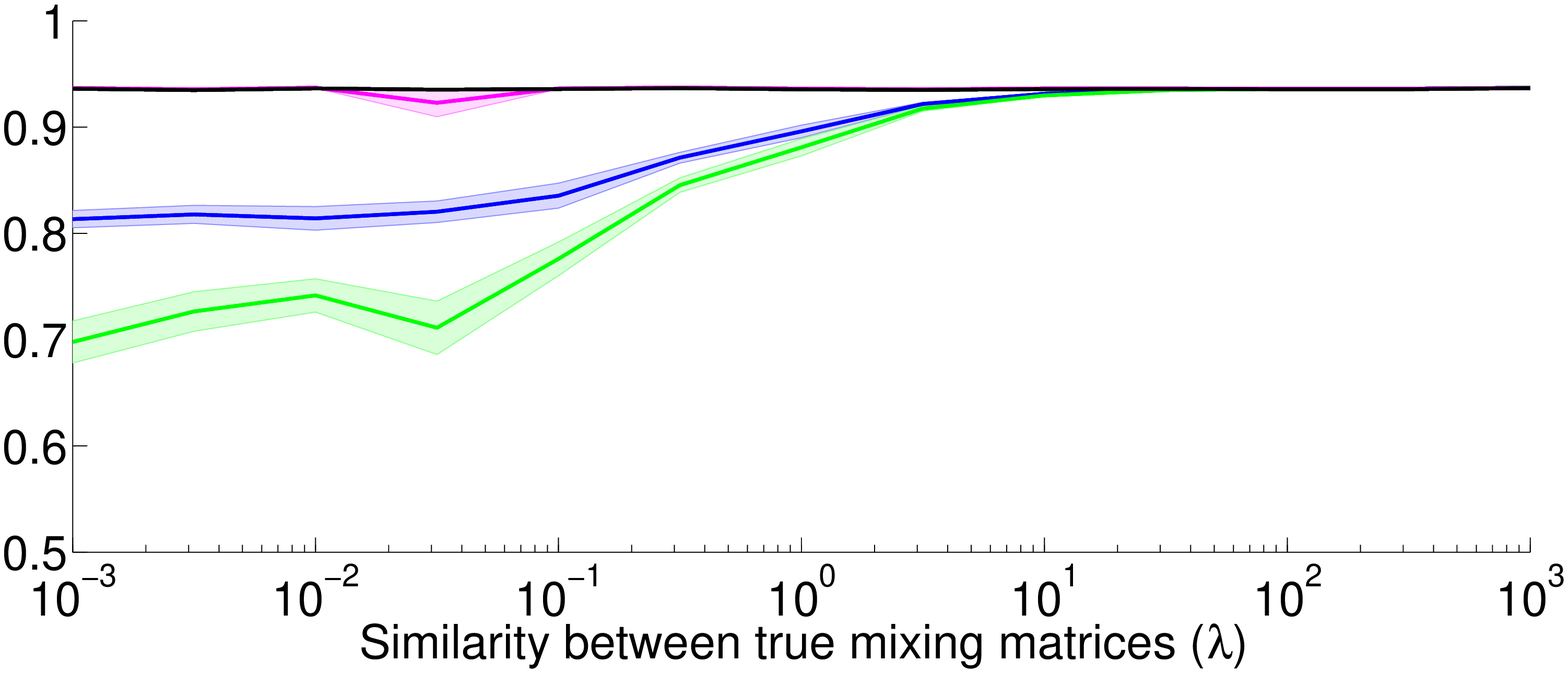}
		\caption{$M=5$}\label{fig.siml}
	\end{subfigure}
	\caption{Performance of BCorrCA, GFA, CorrCA and CCA on simulated data measured by mean correlation coefficient and standard error of the mean with respect to the true source over 20 repetitions. In each subfigure the data is varied in one condition: In \textbf{(a)} and \textbf{(b)} the performance is tested under different levels of SNR and either 2 or 5 views. \textbf{(c)} shows the performance with varying number of views, and in \textbf{(d)} the similarity between the true weights are varied by the $\lambda$ parameter. Both data for both \textbf{(c)} and \textbf{(d)} are created with $\textnormal{SNR}=-6$. In all subfigures $\lambda=10^{-3}$, except for \textbf{(a)} where $\lambda=10^3$.} 
	\label{fig.simres}
\end{figure*}
We report a more extensive set of tests in which the 'true' similarity between the views, the SNR, or the number of views are varied, see figure \ref{fig.simres}. In the following the results will be reviewed one condition at a time. In these tests we used a single  hidden source.

{\noindent \textbf{Signal-to-noise SNR}}\\ 
For the two view case (figure \ref{fig.simsnr1}) and high SNR the algorithms all perform equally well, suggesting that they reach the \reviewed{same} solutions, but as the noise level increases GFA shows a quick drop towards zero correlation. This drop is also be seen in the case of five views in figure \ref{fig.simsnr4}, in this case for both latent variable models and is due to both models choosing an over-regularized zero-source solution as the cost of a poor estimation gets too high. \reviewed{GFA and BCorrCA are both derived using variational inference and share the majority of their priors, the reason for BCorrCA outperforming GFA at low levels of SNR can therefore either be attributed to the introduction of the \textbf{U} and $\lambda$ constraints or the full rank model for the noise.}


{\noindent \textbf{Number of views}}\\
Figure \ref{fig.simm} illustrates how performance improves as the number of available views increase. Note that this increase is not due to an increase in the amount of data, as we have fixed the number of observations. BCorrCA and GFA are build to generalise to arbitrary number of views. Here,  CCA and CorrCA actually perform worse when there are more than two views and the true mixing matrices are different. With identical simulated mixing matrices the algorithms perform at par with BCorrCA. This suggests that the increased performance stems from having more observations of the same signal which effectively averages out the additive noise.

{\noindent \textbf{Similarity between mixing matrices}}\\
All tests were run at different levels of similarity between the true mixing matrices of each dataset as attained by varying the $\lambda$ parameter. Figure \ref{fig.siml} shows how this affects the performance in the case of five views. It can be seen that both CCA and CorrCA are affected in a negative way by dissimilar true mixing matrices, whereas BCorrCA and GFA are unaffected. Figure \ref{fig.simsnr1} shows that CorrCA attains the highest correlations at very low levels of SNR when the mixing matrices are equal, which is the case for both two and more views. BCorrCA attains the same increase in performance at SNR levels below -15 dB. Combining this with the fact that BCorrCA performs well with unequal mixing matrices shows that it is able to adapt to a given dataset and regulate the mixing matrices to be independent as in CCA or equal as in CorrCA.

\subsection{Blind source separation}
\begin{figure}
\centering
	\includegraphics[width=.8\linewidth]{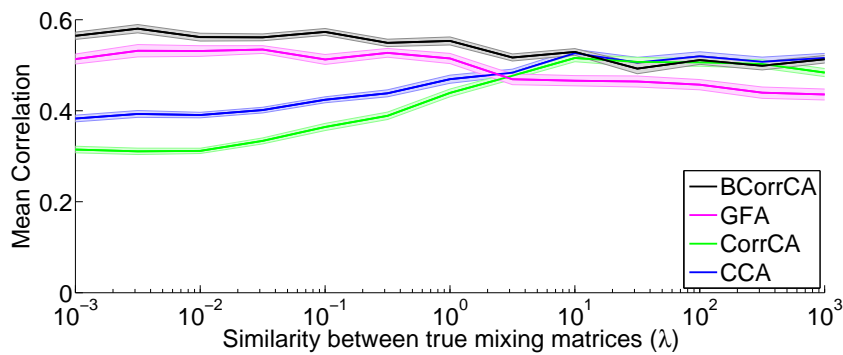}
	\caption{Performance with four hidden sources, five views, $\textnormal{SNR} = -6$dB, and varied similarity between the true weights. Note that the range of the y-axis is lower compared to figure \ref{fig.siml}.}
	\label{fig.K4varL}
\end{figure}
\reviewed{That the difference in performance between BCorrCA and GFA is present when the representations are independent as illustrated, means that at some general difference is due to the full rank noise model. To investigate if modelling inter-view dependencies in the representations also has an effect the difficulty was increased by}
increasing the number of hidden sources to \reviewed{$K_0 = 4$} as used in \cite{Klami2013}\reviewed{. This }decreases the mean correlation, but it does not change the relative performance between the algorithms, when varying SNR or the number of views. A difference can, however, be seen when varying the similarity between the mixing matrices, which is illustrated in figure \ref{fig.K4varL}. \reviewed{Increased similarity can be seen to cause a drop in performance for the two latent variable models, but where the performance of GFA drops below CCA and CorrCA, BCorrCA seem to converge with these two algorithms. Since the noise level is held constant the relative difference in performance can be attributed to the introduction of the \textbf{U} and $\lambda$ constraints.}

The effect from the similarity between mixing matrices seems to have an interaction with the level of SNR, as high similarity increases performance for very low values of SNR (below -10 dB). 
\section{Validation on benchmark EEG from a visual selective attention task}\label{sec:EEG}
For the purpose of evaluating the algorithm on real EEG, we use a well-documented data set from the five-box task study by the Swartz Center for Computational Neuroscience \citep{Makeig1999}.
\subsection{Task design}
ERPs were recorded from subjects attending to 5 laterally arranged boxes 0.8 cm above a central fixation point on a monitor in which a filled circle appeared in random order for a duration of 100 ms. In each block of trials the subject focused on a highlighted box and pressed a thumb response button when a disc appeared in the attended location \citep{Townsend1994}.
\subsection{Pre-processing}
The continuous EEG was recorded at 512 Hz from 60 subjects from 29 channels referenced to the left mastoid and mounted in the international 10-20 system. The EEG was segmented into response-locked epochs with 100 ms pre- and 290 ms post-response time (n=200). Epochs with values higher than 70$\mu V$ or exceeding 5 standard deviations were automatically rejected to remove eye artefacts and electrical drift. The remaining epochs were bandpass filtered using a windowed sinc-filter with passband frequencies 0.1-40 Hz and baseline-corrected in relation to the 100 ms pre-response interval. The "infomax" ICA algorithm \citep{Bell1995} was then used on the concatenated epochs to isolate independent components containing eye artefacts. Scalp topographies and time courses were used to identify noisy components which were then removed from the EEG which was then normalised with respect to its total power.
\subsection{Previous analysis}By analysing this data using ICA, \cite{Makeig1999} were able to decompose the late positive complexes evoked by visual stimuli into a frontoparietal component (P3f), a longer-latency large component (P3b) and a postmotor potential (Pmp) induced by a button press and later \cite{Delorme2007a} found with ICA and time-frequency analysis that faster responses were linked with larger P3f peaks.
\subsection{Finding correlated components}
Because BCorrCA is sensitive to temporal misalignment the criterion for including views \reviewed{(subjects)} in this \reviewed{experiment} was based on the correlation of the averaged ERP of channel Pz between view 12 and all of the views in the cohort. The particular view and channel was chosen as reference due to a strong response and a low noise level. The six views with highest correlation (12,13,14,15,16,22) were chosen and cropped to contain the same number of epochs (343). To test the reliability of the algorithm 100 epochs were randomly drawn from each view and concatenated to create 6 views of "continuous" EEG. \reviewed{This leaves an inference problem of M=6, D=29, N=20000 and K=1 for BCorrCA and K=7 for GFA since the algorithm models noise in components. In GFA the component with the largest variance was chosen.} The procedure was repeated 100 times in order to compute mean and standard error of the mean. Inspired by the approach by \cite{Wipf2010} the prior on the covariance \reviewed{in BCorrCA} is here estimated from the pre-stimulus EEG.
\subsection{Results}
The ratio of the variance of view-specific patterns and the variance of the difference between view-specific patterns represents an estimate of the similarity across views. The same applies to the time series components but the algorithm only computes a common component for all the views. However, by constructing the backward model filters, \textbf{W}, the view-specific components can be found by $\mathbf{y}^{(m)} = {\mathbf{X}^{(m)}}^\intercal \mathbf{w}^{(m)}$. Due to the model formulation $\overset{\sim}{\mathbf{X}} = \textbf{Az}^\intercal$, the patterns and components can only be inferred up to a scale factor, i.e. $  \overset{\sim}{\mathbf{X}} = \textbf{Az}^\intercal = \tfrac{1}{c}\textbf{A} \cdot c \textbf{z}^\intercal$. Fair comparison between the methods then requires normalisation of \textbf{A} \reviewed{by multiplication of} the standard deviation of \textbf{z} and vice versa, when comparing the respective variables. To increase readability the variables are further standardised with respect to the standard deviation of the grand average pattern or ERP. The average variance within views for both patterns and time series is computed as the usual variance

\begin{align}
var_{within}(\boldsymbol{\delta}) = \frac{1}{M}\sum_m^M E[(\delta_m - E[\delta_m])^2] \label{eq:within_variace}
\end{align}

and the between-view variance as the average of the upper triangular sans diagonal of pairwise differences

\begin{align}
var_{between}(\boldsymbol{\delta}) = \frac{2}{M(M-1)}\sum_{m=1}^M\sum_{i=m+1}^M E[(\delta_{mi} - E[\delta_{mi}])^2], \quad \delta_{mi} = \delta_m - \delta_i \label{eq:between_variace},
\end{align}

where \(E[\cdot]\) is either the spatial or temporal mean depending on usage. In table \ref{tab:variance} we see that the average within-view variance for both patterns and time series are larger than the average between-view variance which suggests a high amount of universality in the neural representations.  The shared representation is particularly evident from the comparison of time series but as the selection of views were based on ERP similarity this ratio is biased in favour of low variation between views.

\begin{figure}
\begin{center}
	\includegraphics[width=0.5\linewidth]{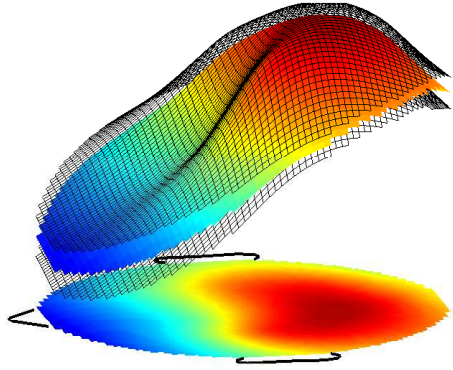}
\end{center}
 \caption{Grand average pattern projection and standard error (mesh) attained with BCorrCA. The pattern is illustrated both as a 2- and 3-dimensional structure to show the between-view variation. Positive weights are located in the parietal region.}\label{fig:filter_projection_3D}
\end{figure}

\begin{figure}
\begin{center}
	\includegraphics[width=0.5\linewidth]{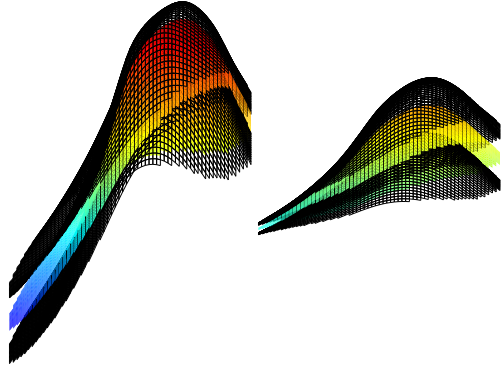}
\end{center}
 \caption{Grand average pattern projection and standard error (mesh) for BCorrCA (left) and GFA (right). The patterns are normalised wrt. the standard deviation of the component \textbf{z}. The patterns are illustrated as 3-dimensional structures to show the difference in amplitude of the weights. Positive weights are located in the parietal region. \reviewed{The left figure is the same as the 3D part of figure \ref{fig:filter_projection_3D}.}}\label{fig:five_box_filter_projection_3D_comparison}
\end{figure}

In figure \ref{fig:filter_projection_3D} the evidence for a shared representation is \reviewed{seen in} the relatively small standard deviation (mesh) compared to the average pattern. The scalp topography shows positive parietal and negative frontal activation which is very close to the pattern found to isolate the Pmp in the late positive complex \citep{Makeig1999}.  In figure \ref{fig:five_box_filter_projection_3D_comparison} we see how BCorrCA is able to acquire a numerically stronger representation but maintain roughly the same standard deviation on an electrode basis between subjects.

\begin{figure}
\begin{center}
\includegraphics[width=.8\linewidth]{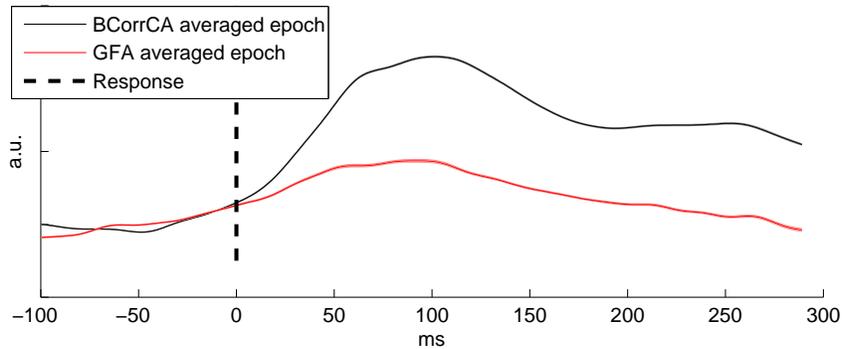}
\end{center}
\caption{Average ERP of component from BCorrCA and GFA. Mean and standard error is computed from 100 repetitions of 100 randomly selected epochs from each view. The small error indicates a high level of reliability for both methods. The epochs are normalised wrt.\ to the mean standard deviation of the mixing matrices, \textbf{A}. \reviewed{Figure \ref{fig:five_box_filter_projection_3D_comparison} and \ref{fig:avg_comp_epoch} show that we have higher variance explained, by scaling $var(z) = 1$, respectively $var(A) = 1$.}}\label{fig:avg_comp_epoch}
\end{figure}

Figure \ref{fig:avg_comp_epoch} shows in the small error a high level of reliability for BCorrCA and GFA. The neural response is represented by a stronger signal in BCorrCA, which is due to the difference in pattern weights as we saw in figure \ref{fig:five_box_filter_projection_3D_comparison}, and is closer to the response we expected to see from \cite{Makeig1999}, suggesting BCorrCA has greater affinity to physiologically meaningful solutions.

\reviewed{
To further quantify the performance of each method the proportion of variance explained (PVE) was calculated from the reconstructed data by 
\begin{align}
\text{PVE} = \frac{\text{SST} - \text{SSE}}{\text{SST}},
\end{align}
where SST is the total sum of squares and SSE the error sum of squares, ie. the residual variance. The proportion of variance explained across all views (subjects) and repetitions is very close for the two methods but a little higher for GFA. However, the variation of PVE is also much higher and a closer analysis disclosed a significant difference in PVE between repetitions for GFA and less so for BCorrCA. In most repetitions GFA finds a very good solution for a single view and poor solutions for the rest whereas BCorrCA is more steady across repetitions and attains only mediocre PVE results for all views. GFA thus has a tendency to 'overfit' to a single view which may explain why, when averaged over multiple repetitions, it attains results with a lower variance, however, using the average has the advantage of obtaining a solution with a more common neural pattern. The latent 'response component' is very consistent across views due to its low-level cognitive origins so even though it is explained mostly by a single view, it is a good representation of the shared component. Since GFA has no constraints on the patterns this solution is satisfactory, and perhaps preferred, in a context where a strong shared pattern is not part of the hypothesis. 
}

\begin{table}
\begin{center}
\renewcommand{\arraystretch}{1.3}
\caption{\small Mean variance of time series components and patterns calculated according to \eqref{eq:within_variace} and \eqref{eq:between_variace} with standard error of the mean variance in parentheses. For comparability patterns and time series are normalised wrt. the standard deviation of the grand average for each method.}
\label{tab:variance}
	\begin{tabular}[C]{rcccc}
	& \multicolumn{2}{c}{\textbf{BCorrCA}} & \multicolumn{2}{c}{\textbf{GFA}} \\
	& Within-view & Between-view & Within-view & Between-view\\\hline 
	Patterns & 1.39 (0.20) & 0.93 (0.20) & 1.47 (0.29)& 1.13 (0.23)\\
	Time series & 1.09 (0.06) & 0.23 (0.05) & 1.08 (0.17)& 0.21 (0.03)
	\end{tabular}	
\end{center}
\end{table}

\section{Discussion and Conclusion}
Research in social neuroscience has shifted from being inherently single person studies of people observing others towards multi-way interaction between multiple persons \citep{Schilbach2013a}, which calls for methods that are able to adapt to the level of universality in neural representations across brains. The probabilistic implementation of correlated component analysis presented here provides a new approach to the extraction of flexible shared representations and information.

Extensive tests in simulated data show that the proposed model and inference scheme is able to estimate the similarity between the views and that an increasing number of views in fact improves the estimate. By adjusting the level of additive observation noise we demonstrated that BCorrCA performs better than GFA in poor signal to noise conditions and better than all the other methods when the number of views increases and there is limited similarity between mixing patterns.
The effect of increasing the number of views is positive for both BCorrCA and GFA, while CorrCA and CCA stabilise at a lower level of performance (similarity between estimated sources and 'true' sources), when the patterns are dissimilar. However, when the patterns become similar they all reach the same level of source estimation.
Our experiments in simulated data also showed that multiple views improve the extraction of shared signals, even when the total amount of observations were kept the same. By
 these simulations we thus conclude that the proposed method is better suited for multi-view problems than CorrCA and CCA, and to be preferred over GFA when the SNR is low. And indeed offers the flexibility to infer the degree of universality in the view representations.

Both of the latent variable models were seen to have a drop in performance when the data contained multiple sources and their mixing patterns were similar across views. Inspection of  the inferred sources show that they appear to be linear combinations of the true sources, which suggests that the latent variable models may experience a rotation of the 'true' source space. Though a solution to the rotation problem was discussed in \cite{Klami2013}, further investigation into this issue is warranted.


\reviewed{
Latent variable models usually use the lower bound for the full log marginal likelihood to monitor convergence, but it in some cases it is also used to choose between multiple analyses on the same dataset \citep{Klami2013}. This prompted an investigation into the relationship of the lower bound and the correlation between the true sources and the inferred ones. Though some dataset-specific variation was seen, the average result was a correlation close to zero. The reason for the lack of a consistent correlation between the lower bound and the mean true source correlation could stem from the fact that the lower bound represents how close the entire model is to the observations. This could mean that an analysis can achieve a low mean correlation with the true sources, but still model the observations correctly through the other variables, e.g., noise variables, and thus obtain a high lower bound. The tests, however, showed a strong correlation between the lower bound and the number of active sources, meaning that the analysis with highest lower bound would likely be the one with the correct number of active sources. Certain types of data caused GFA to produce results with a lower number of active sources than the true number of sources, and in roughly a third of the datasets did these analyses obtain a higher lower bound than the ones obtaining the correct number of active sources. While the lower bound in all but some special cases seems to be a valid unsupervised performance measure for selecting the solution with the correct the number of active sources, further investigations into this area, e.g., comparing the lower bound with other performance measures, might be able to give a clearer understanding of its proper use.}

Our analysis of the visual selective attention shift EEG dataset showed the expected post-motor potential response following the key press is markedly stronger using BCorrCA compared to GFA and thus explains more of the variation. In Table I we list the total variation within and between subjects (views). Note, the within subject variability (i.e., variability across electrode weights) is larger than the between subject variability, adding evidence to the hypothesis that subjects to some degree share spatial representation. The estimated response time course is highly universal for the subjects entering the analysis, as expected. \reviewed{We found that while GFA initially seemed to produce inferior results the issue was caused by solutions 'overfitting' to individual views. The evaluation criteria in this test punish this behaviour while it may provide a better solution in another setting.}

The proposed scheme for inference of shared responses in the face of inter-subject representational variability thus offers a new analytic route for experiments with simultaneous stimulation of groups. Rather than analysing a large number of pairwise correlations between of subject responses the proposed scheme enable inference of joint attention and other joint activity in large cohorts.

\subsubsection*{Acknowledgments}
We thank the reviewers many constructive comments that helped improve the manuscript. This work is supported by Innovation Fund Denmark project Neuro24/7.

\bibliographystyle{iclr2017_conference}
\bibliography{library}

\appendix
\section*{Appendix}
\subsection*{Note on change in performance on simulated data}
It should be noted that the performance of BCorrCA has improved with low levels of SNR (particularly in the two view case) compared to an earlier article regarding BCorrCA \citep{Poulsen2014}. This was due to an error in the lower bound calculations, which gave a wrong estimate of the time of convergence, and caused the algorithm stop its iterations prematurely. A change in the data creation with regards to adding observation noise with the correct amplitude and avoid differences in scaling has caused an improvement in the performance of all four algorithms, but has not altered their relative performance.

\section*{Correlated components as an eigenvalue problem}
To introduce the notion of correlated components, we briefly review the multivariate schemes CCA and CorrCA, both of which are valid for two views.
Given two multivariate spatio-temporal views, $\textbf{X}^{(1)} \in \mathbb{R}^{D_1\times N}$ and $\textbf{X}^{(2)} \in \mathbb{R}^{D_2\times N}$, with $\{D_1,D_2\}$ being the number of measured features in the views and $N$ the number of time samples, CCA estimates weights, $\{\textbf{W}^{(1)},\textbf{W}^{(2)}\}$, which maximise the correlation between $\textbf{y}_1 = {\textbf{X}^{(1)}}^\intercal\textbf{w}_k^{(1)}$ and $\textbf{y}_2 = {\textbf{X}^{(2)}}^\intercal\textbf{w}_k^{(2)}$. At the same time CCA constrains the estimated weights with the condition that ${\textbf{X}^{(1)}}^\intercal\textbf{w}_k^{(1)}$ and ${\textbf{X}^{(1)}}^\intercal\textbf{w}_{k'}^{(1)}$ are uncorrelated for $k\neq k'$ \cite{Klami2013}. Introducing the sample covariance matrix, $\textbf{R}_{ij}=\frac{1}{N}\textbf{X}^{(i)}{\textbf{X}^{(j)}}^\intercal$, CCA estimates the weights analytically through eigenvalue decompositions

\eq{
\textbf{R}_{11}^{-1}\textbf{R}_{12}\textbf{R}_{22}^{-1}\textbf{R}_{21}\textbf{w}^{(1)} &= \rho \textbf{w}^{(1)} \\
\textbf{R}_{22}^{-1}\textbf{R}_{21}\textbf{R}_{11}^{-1}\textbf{R}_{12}\textbf{w}^{(2)} &= \rho \textbf{w}^{(2)} \notag.
}

Correlated component analysis (CorrCA) is a related approach for the case where the views are similar, i.e., $D_1 = D_2$, for which it imposes the additional constraint of shared weights $\textbf{w} = \textbf{w}^{(1)} = \textbf{w}^{(2)}$. This stronger universality assumption can potentially increase sensitivity as it involves estimation of fewer parameters.  In correlated component analysis the weights are thus estimated through a single eigenvalue problem \citep{Dmochowski2012},

\eq{
\left(\textbf{R}_{11} + \textbf{R}_{22}\right)&^{-1} \left(\textbf{R}_{12} + \textbf{R}_{21}\right)\textbf{w} = \rho \textbf{w}.\label{eq.correlated component analysis}
}

\subsection*{Robustness of CorrCA with differently mixed views} \label{subsec.cocacap}
By definition CorrCA could be challenged, if the 'true' weights of the views were different. However, in initial tests on simulated data we found only a small drop in performance (data not shown). To understand this robustness we analyse a kind of 'worst case scenario', in which the 'true' view specific mixing weights are orthogonal.

The observations are assumed to consist of a single 'true' signal mixed into $D$ dimensions with additive Gaussian noise; $\textbf{X}^{(1)} = \textbf{a}^{(1)}\textbf{z}^\intercal + \boldsymbol{\epsilon}\ $, $\textbf{X}^{(2)} = \textbf{a}^{(2)}\textbf{z}^\intercal + \boldsymbol{\epsilon}\ $. Given a large sample, the covariance matrices are given as $\textbf{R}_{11} = P\cdot\textbf{a}^{(1)}{\textbf{a}^{(1)}}^\intercal  + \sigma^2 \textbf{I}\ $, $\quad \textbf{R}_{12} = P\cdot\textbf{a}^{(1)}{\textbf{a}^{(2)}}^\intercal$, where $P$ is the variance of \textbf{z} and $\sigma^2$ signifies the noise variance. For simplicity the weight vectors are assumed to be unit length. The two matrices in (\ref{eq.correlated component analysis}) can then be written as

\begin{align}
(\textbf{R}_{11} + \textbf{R}_{22})^{-1} &= \frac{1}{P}\left([\textbf{a}^{(1)}\ \textbf{a}^{(2)}]  \wvec  + \frac{2\sigma^2}{P} \textbf{I}\right)^{-1}  \label{eq.r11r22}\\
\textbf{R}_{12} + \textbf{R}_{21} &= P\cdot [\textbf{a}^{(1)}\ \textbf{a}^{(2)}]  \wvectwo , \label{eq.r12r21}
\end{align}
using block matrix notation.  Using ${\textbf{a}^{(1)}}^\intercal\textbf{a}^{(2)} = 0$, $\|\textbf{a}^{(1)}\|^2 = \|\textbf{a}^{(2)}\|^2 = 1$ and the Woodbury identity, the matrix product of (\ref{eq.r11r22}) and (\ref{eq.r12r21}) can be expressed as

\begin{align}
\left(\textbf{R}_{11} + \textbf{R}_{22}\right)^{-1} \left(\textbf{R}_{12} + \textbf{R}_{21}\right) &= \frac{P}{2\sigma^2}\left( \textbf{I} - \frac{P}{2\sigma^2 + P}[\textbf{a}^{(1)}\ \textbf{a}^{(2)}] \wvec \right)[\textbf{a}^{(1)}\ \textbf{a}^{(2)}]  \wvectwo \lb
%
&= \frac{P}{2\sigma^2 + P}(\textbf{a}^{(1)}{\textbf{a}^{(2)}}^\intercal + \textbf{a}^{(2)}{\textbf{a}^{(1)}}^\intercal ) \label{eq.cocamatrix}
\end{align}
Using the simplifying assumptions made earlier, an eigenvector for (\ref{eq.cocamatrix}) can be seen to have the form $\alpha\textbf{a}^{(1)} + \beta\textbf{a}^{(2)}$ since

\eq{
\frac{P}{2\sigma^2 + P}(\textbf{a}^{(1)}{\textbf{a}^{(2)}}^\intercal + \textbf{a}^{(2)}{\textbf{a}^{(1)}}^\intercal ) &(\alpha\textbf{a}^{(1)} + \beta\textbf{a}^{(2)}) = \frac{P}{2\sigma^2 + P}(\alpha\textbf{a}^{(2)} + \beta\textbf{a}^{(1)}).
}
It can be seen that $\alpha\textbf{a}^{(1)} + \beta\textbf{a}^{(2)}$ is an eigenvector when either $\alpha = \beta$ or $\alpha = -\beta$ with $\pm\frac{P}{2\sigma^2 + P}$ as eigenvalues. This means that when the true mixing weights of two views are orthogonal CorrCA finds a useful common set of weights, consisting of a weighted sum of the true weights, leading only to a limited drop in SNR. 

\subsection*{Updates for BCorrCA} \label{app.updates}
Here follows the updates for each variable in BCorrCA, which are iterated in an expectation-maximisation like manner until convergence.

\eq{
q(\textbf{Z}) &= \prod_{n=1}^N \mathcal{N}\left( \textbf{z}_n|\boldsymbol\mu_{\textbf{z},n},\Sigma_\textbf{z} \right) \\
\Sigma_\textbf{z}^{-1} &= \sum_m^M\left\{\mom{{\textbf{A}^{(m)}}^\intercal\Psi^{(m)}\textbf{A}^{(m)}}\right\}+\textbf{I} \\
\boldsymbol\mu_{\textbf{z},n} &= \Sigma_\textbf{z}\sum_m^M\mom{{\textbf{A}^{(m)}}^\intercal}\mom{\Psi^{(m)}}\textbf{x}_n^{(m)}\lb
q(\boldsymbol\Psi) &= \prod_{m}^M \mathcal{W}\left( \textbf{S}_\Psi^{(m)},v_\Psi \right) \\
{\textbf{S}_\Psi^{(m)}}^{-1} &= \mom{\textbf{A}^{(m)}\sum_{n}^N \textbf{z}_n\textbf{z}_n^\intercal{\textbf{A}^{(m)}}^\intercal} + \sum_{n}^N \textbf{x}_n^{(m)}{\textbf{x}_n^{(m)}}^\intercal \notag\\
& \qquad\quad - 2\cdot\sum_{n}^N\textbf{x}_n^{(m)}\mom{\textbf{z}_n^\intercal}\mom{{\textbf{A}^{(m)}}^\intercal} + \textbf{S}_0^{-1} \\
v_\Psi &= N+v_0 
}\eq{
q(\textbf{A}^{(m)}) &= \prod_{d=1}^D \mathcal{N}\left( \hat{\textbf{a}}^{(m)}_d|\ \boldsymbol\mu^{(m)}_{\textbf{a}_d} ,\Sigma^{(m)}_{\textbf{a}_d} \right) \\
{\Sigma^{(m)}_{\textbf{a}_d}}^{-1} &= \mom{\psi^{(m)}_{dd}}\sum_{n}^N\mom{\textbf{z}_n\textbf{z}_n^\intercal} + \mom{\lambda}\textbf{I} \\
\boldsymbol\mu^{(m)}_{\textbf{a}_d} &= \Sigma^{(m)}_{\textbf{a}_d}\left( \sum_{n}^N\mom{\textbf{z}_n}\mom{\psi^{(m)}_{(d,:)}}\textbf{x}_n^{(m)} + \mom{\lambda}\mom{\textbf{u}_{d}} \right. \notag\\
  &\quad \left. -\sum_{d'\neq d}^D \mom{\psi^{(m)}_{dd'}}\sum_{n}^N\mom{\textbf{z}_n\textbf{z}_n^\intercal}\mom{{\textbf{a}_{d'}^{(m)}}^\intercal}  \right) 
}\eq{
	q(\textbf{U}) &= \prod_{k=1}^K \mathcal{N}\left( \textbf{u}_k|\ \boldsymbol\mu_{\textbf{u}_k} ,\sigma^{2}_{\textbf{u}_k}\textbf{I} \right) \\
{\sigma^{-2}_{\textbf{u}_k}} &= M\mom{\lambda} + \mom{\alpha_k}\\ 
\boldsymbol\mu_{\textbf{u}_k} &= {\sigma^{2}_{\textbf{u}_k}}\mom{\lambda} \sum^M_m\mom{\textbf{a}^{(m)}_{k}} 
}\eq{
q(\boldsymbol\alpha) &= \prod_{k}^K\mathcal{G}a(\alpha_k|a_\alpha,b_{\alpha_k})\\
a_\alpha &= a_0 + \frac{D}{2} \\
b_{\alpha_k} &= b_0+\frac{\mom{\textbf{u}^{\intercal}_k\textbf{u}_k}}{2} \lb
q(\lambda) &= \mathcal{G}a(\lambda|a_\lambda,b_\lambda)\\
a_\lambda &= a_0+\frac{MKD}{2} \\
b_{\lambda} &= b_0 + \sum^K_{k} M\frac{\mom{\textbf{u}^{\intercal}_k\textbf{u}_k}}{2} + \sum^M_m \left\{ \frac{\mom{{\textbf{a}_k^{(m)}}^\intercal\textbf{a}^{(m)}_k}}{2} \right. \notag\\
&\quad \left.- \mom{{\textbf{a}_k^{(m)}}^\intercal}\mom{\textbf{u}_k}\right\}.
}
where $\hat{\textbf{a}}^{(1)}_d$ is a column vector corresponding to the $d$'th row of \textbf{A} and $\mom{.}$ signifies the expectation. Note that $v_\Psi$, $a_\alpha$ and $a_\lambda$ are constants and can be defined before iterating over the other updates.

\subsection*{Variational Message Passing}
The factorisation invoked in variational inference can be viewed as the decomposition of a large network into a subset of factors that individually can be approximated variationally, creating a message--passing like algorithm. Variational message passing (VMP) by \cite{Winn2005} is an efficient implementation of this principle in which each factor is only conditioned on variables in the same Markov blanket. Because VMP constrains the factors to be in the exponential family and conjugate with respect to the distributions they are conditioned on (their parents), the variational updates simplify greatly \citep{Attias2000}. It is thus possible to write a conditional distribution of the exponential family on a generic form that allows the algorithm to extract sufficient statistics and pass on as a message. The receiving node can then update its posterior belief based on  the incoming messages.

VMP is a special case of a broader set of message--passing algorithms that all rely on minimising the $ \alpha $--divergence. What makes VMP unique is that it, like mean field theory, seeks the minimisation of the exclusive KL--divergence ($ \alpha=0 $), which ensures that minimising the local divergence exactly minimises the global divergence \citep{Minka2005}. For  VMP inference we use the probabilistic programming framework, infer.NET, developed  by \cite{inferNet}.

\begin{figure}
\centering
	\includegraphics[width=\linewidth]{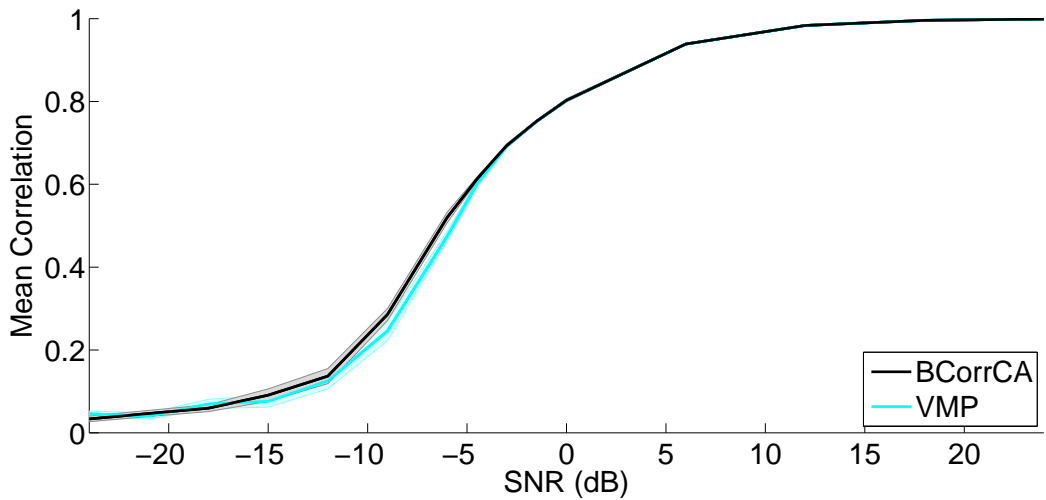}
	\caption{Comparison at varying levels of SNR between BCorrCA implemented in Matlab using the derivations presented in this article, and a version implemented with VMP using Infer.NET . $\lambda$ has been set equal to $10^3$ making the true weights of both views nearly equal. The shown correlation coefficient is calculated as the mean of 20 simulations at each condition, with the standard error of the mean illustrated as the opaque area.}
	\label{fig.vmp}
\end{figure}

Figure \ref{fig.vmp} shows the performance of BCorrCA implemented in Matlab using the derivations presented in this paper and a version implemented with VMP at multiple levels of SNR. For each of the 20 repetitions the algorithms were initialised identically. Considering their standard error of mean, the two implementations of BCorrCA were deemed to have equal performance, with the difference being attributed to differences in the order of updates. Therefore the VMP implementation was left out of the rest of the tests presented in this article.

\end{document}